\documentclass[conference]{IEEEtran}
\IEEEoverridecommandlockouts
\usepackage{cite}
\usepackage{amsmath,amssymb,amsfonts}
\usepackage{algorithmic}
\usepackage{graphicx}
\usepackage{textcomp}
\usepackage{xcolor}
\def\BibTeX{{\rm B\kern-.05em{\sc i\kern-.025em b}\kern-.08em
    T\kern-.1667em\lower.7ex\hbox{E}\kern-.125emX}}
\usepackage{hyperref}
\usepackage{url}
\usepackage{dirtytalk}
\renewcommand{\vec}[1]{\mathbf{#1}}
\usepackage{bbm}
\usepackage{amsmath}
\usepackage{amsthm}
\usepackage{graphicx}
\usepackage{color}
\usepackage{soul}
\usepackage{cite}
\usepackage[caption=false,font=footnotesize]{subfig}
\usepackage{enumerate} 
\usepackage{hyphenat}
\usepackage{amssymb}
\usepackage{textcomp}
\usepackage{xcolor} 
\begin{document}

\title{Unknown Delay for Adversarial Bandit Setting with Multiple Play\\
}

\author{\IEEEauthorblockN{Olusola T. Odeyomi}
\IEEEauthorblockA{\textit{Department of Electrical Engineering and Computer Science} \\
\textit{Wichita State University}\\
Wichita, Kansas, United States \\
otodeyomi@shockers.wichita.edu}

}

\maketitle

\begin{abstract}
This paper addresses the problem of unknown delays in adversarial multi-armed bandit (MAB) with multiple play. Existing work on similar game setting focused on only the case  where the learner selects an arm in each round. However, there are lots of applications in robotics where a learner needs to select more than one arm per round. It is therefore worthwhile to investigate the effect of delay when multiple arms are chosen. The multiple arms chosen per round in this setting are such that they experience the same amount of delay. There can be an aggregation of feedback losses from different combinations of arms selected at different rounds, and the learner is faced with the challenge of associating the feedback losses to the arms producing them. To address this problem, this paper proposes a delayed exponential, exploitation and exploration for multiple play (DEXP3.M) algorithm. 
The regret bound is only 
slightly worse than the regret of DEXP3 already proposed for the single play setting with unknown delay.     
\end{abstract}

\section{Introduction}
Online learning is a very powerful theoretical framework for studying repeated games where a learner makes some predictions on some arbitrary sequence of loss (or reward) functions generated either from a fixed but unknown probability distribution or by an adversary \cite{shalev2012online,blum1998line,fiat1998online}. The learner in practice selects one or more arm(s) of unknown quality and faces the trade-off between exploiting profitable past arm(s) and exploring new arm(s) with which the learner has little or no information. Online learning studied as a multi-armed bandit (MAB) problem is classified according to how losses (or rewards) are generated. In stochastic multi-armed bandit, losses are generated from a fixed but unknown distribution \cite{bubeck2012regret,auer2010ucb,audibert2007tuning,dani2008stochastic}, while in non-stochastic or adversarial multi-armed bandit, losses are generated arbitrarily to deceive the learner such as in game-theoretic settings \cite{auer1995gambling,bubeck2012regret}. These classic MAB variants have been well studied for the single play setting where the learner chooses an arm per round and observes the loss of the arm chosen at the end of each round, and the multiple play setting where the learner chooses many arms per round and observes the losses of the chosen arms at the end of each round \cite{xia2016budgeted,uchiya2010algorithms,zhou2018budget}.

A tougher setting than the classic MAB setting is when the feedback of the chosen arm undergoes some delays, and it is not observed at the end of each round. There is a good number of studies on the stochastic MAB single play setting with delays. These delays are  classified into various forms such as fixed delays \cite{dudik2011efficient}, where the delays have a common fixed unknown value; non-anonymous random delays \cite{joulani2013online,mandel2015queue}, where the delays are random but the feedback losses and the arms producing those losses are known; and aggregated anonymous delays \cite{pike2017bandits}, where only the sum of the feedback losses over some unknown rounds of play and over some unknown chosen arms  is known. There is also the conversion delay \cite{vernade2017stochastic} applicable in web advertisement but has the limitation that the delay distribution must be known; and stochastic bounded delays in Gaussian process bandit \cite{desautels2014parallelizing}.

In the adversarial setting, there has also been a good number of research work done for the single play setting with delays. These delays are also classified into various forms. A generalized regret bound on any  \textit{base} MAB algorithm  for adversarial bandit with fixed delays is obtained in \cite{joulani2013online}. In \cite{cesa2019delay}, the authors found a tight regret bound on EXP3 algorithm with fixed delays. For the case of composite anonymous delays \cite{cesa2018nonstochastic}, where the received feedback loss is the sum of losses of some previously played arms over some unknown rounds of play, the authors found a  generalized regret bound on any \textit{base} MAB algorithm. In \cite{li2019bandit}, the authors found a tight regret bound for EXP3 with composite delayed feedback. In the multiple play setting with delays, both the stochastic bandit and adversarial bandit are yet to be studied.  However, in this paper, we study adversarial MAB for the multiple play setting with delays because it is a tougher and a more practical setting. This paper therefore partially bridges the gap and advances knowledge in this area.

To motivate this research problem, it is important to give some practical applications of the problem setting. Consider a  communication network where a transmitter hops across a subset of channels for transmitting some packets  to a receiver each hop time. Due to environmental factors, the receiver receives the packets with delay. The transmitter needs feedback from the receiver each hop time  in order to know if the channels it hops across give good reward based on the quality of the channel. If the feedback is returned at a later time, after the transmitter must have hopped across other subsets of channels, then it becomes difficult for the transmitter to correctly match the subsets of channels to the feedback it receives and learn from it. This problem is also common in online advertisement where an ad company displays divers subsets of ads to stimulate interest in their product to visitors of the website. Each subset of ads is designed to satisfy a different class of visitors. A visitor may pick interest in the product by observing a particular subset of displayed ad at a particular time but he chooses to buy the product at a different time. Therefore, the ad company does not know which subset of ads led to the purchase of the product. Other applications can be found in the medical fields, e-commerce, social media etc.

The research contributions in this paper are: (i) to formulate the setting for adversarial multiple play multi-armed bandit with unknown delays (ii) to develop an efficient algorithm which is named DEXP3.M (iii) to provide a tight theoretical upper bound on the regret.

\subsection{Related work}

Most of the algorithms proposed for single play adversarial MAB with delays are black-box algorithms that wrap around classic adversarial MAB algorithms, called \textit{Base} MABs \cite{cesa2018nonstochastic,joulani2013online}. The settings for these wrapper algorithms are too generic, and their regret bounds are not always tight. Authors in \cite{li2019bandit} followed a different approach by investigating the effects of unknown delays on the famous EXP3 algorithm and found a tighter bound on the regret. This paper therefore follows the same approach in \cite{li2019bandit} to obtain the regret bound on EXP3.M algorithm proposed in \cite{uchiya2010algorithms} for the multiple play setting. This paper is the first of its kind to study the multiple play adversarial setting with unknown delays.

\section{Problem setting}
A review of the classic adversarial MAB with single play is essential before delving into the delayed version.
\subsection{Classic Adversarial Bandit with Single Play}
In the classic adversarial bandit with single play, the learner chooses an arm $i$ at time $s$  from a finite set of arms $\mathcal{A}=\{1,...,i,...K\}$, based on the probability distribution $\vec{p}_s\in\Delta_K$ over all arms. The probability simplex $\Delta_K$ is defined as $\Delta_K:=\{\vec{p}\in \mathbbm{R}_+^K:p(i)\geq 0, \forall i; \sum^K_{i=1}p(i)=1\}$ \cite{li2019bandit}. 
The learner observes the loss $l_s(i_s)$ of only the arm $i_s$  he chooses at time $s$. All other losses are not revealed to the learner. Since there is no delay in feedback, the losses of all chosen arms are known at the end of each timeslot, and the learner can update the probability $\Vec{p}_{s+1}$ from the knowledge of all previous losses $\{l_t(i_t)\}^s_{t=1}$. A randomized algorithm is then developed to help the learner minimize his \textit{regret}. The regret is the difference between the performance of the algorithm and that of a single fixed policy, over the total timeslot $T$. The regret for the classic setting is defined mathematically as

\begin{equation}
 Reg^{C}_{T}:=\sum^T_{s=1}\mathbbm{E}[\Vec{p}^\mathbbm{T}_s\Vec{l}_s]-\sum^T_{s=1}(\Vec{p}^*)^\mathbb{T}\Vec{l}_s,  
\end{equation}
where $C$ means classic, $\mathbb{T}$ means transpose and $\vec{l}_s$ is the $K\times 1$ loss vector containing the incurred loss $l_s(i_s)$ of arm $i$. The expectation is taken over the randomness of $\Vec{p}_s$ that is caused by the random choice of arms $i_t$, chosen from time $t=1$ to time $t=s-1$. The best fixed policy $\Vec{p}^*$ is defined as 

\begin{equation*}
  \Vec{p}^*:=\arg\min_{\vec{p}\in\Delta_K}\sum^T_{s=1}\vec{p}^\mathbbm{T}\Vec{l}_s
\end{equation*}\color{black}

\subsection{Delayed Adversarial bandit with Multiple Play}
In the delayed MAB setting with multiple play, the learner chooses $k$ distinct arms from $\mathcal{A}:=\{1,...,i,...K\}$. The combined $k$ arms chosen at round $s\in \{1,\cdots,T\}$ is called an action $\vec{a}_s$, which is a $k\times 1$  vector component.  Action $\vec{a}_s$ is chosen from the action space $\Vec{C}(\mathcal{A},k):=\{\Vec{a}_s: \text{dim}(\Vec{a}_s)=k\}$. The action space is a compact closed convex set. The cardinality of $\Vec{C}(\mathcal{A},k)$ denotes the possible number of actions from which the learner can choose. This is given as $|\Vec{C}(\mathcal{A},k)|={K \choose k}$. To account for the multiple play setting, the probability simplex $\Delta_K$ is set as the convex hull of $\{\Vec{k}_{\Vec{a}_s}\in \mathbbm{R}^K: \Vec{a}_s\in\Vec{C}(\mathcal{A},k)\}$, where $\Vec{k}_{\Vec{a}_s}$ is a vector whose $j-th$ component is $1/k$, if and only if $j$ is also a component of $\Vec{a}_s$ (Section 4, \cite{uchiya2010algorithms}). From Krein Milman theorem \cite{krein1940extreme}, $\{\Vec{k}_{\Vec{a}_s}\in \mathbbm{R}^K:\Vec{a}_s\in \Vec{C}(\mathcal{A},k)\}$ is a compact convex subset of $\mathbbm{R}^K$ and its set of extreme points is $\Delta_K$. The action $\Vec{a}_s$ should be chosen in an efficient way such that each arm $i\in\vec{a}_s$ is selected with probability $p_s(i)$ . This ensures that $k\mathbbm{E}(\Vec{k}_{\Vec{a}_s})=\Vec{p}_s$. Due to unknown random delays, the loss vector $\l_s(\vec{a}_s)$ is received for action $\vec{a}_s$ after some delays of $d_s$ slots; namely at the end of slot $s+d_s=t$, where $d_s\geq 0$ can vary from slot to slot. The delays $\{d_s\}^T_{s=1}$ can be chosen adversarially. In this paper, we represent the loss vector of an action $\vec{a}_s$, chosen at timeslot $s$ but observed at timeslot $t$ as $l_{s|t}(\vec{a}_{s|t})$. It is assumed that all $i\in \vec{a}_{s|t}$ arms chosen in timeslot $s$ undergo same delay $d_s$. It is possible for the order of feedback to be arbitrary such that $s+d_{s}\geq z+d_{z}$ when $s<z$ and $s,z\in\{1,...,T\}$. Since the delays are arbitrary, it is equally possible for the feedback of many actions, selected in the past at different times to be received at the same time. We denote the set of all such feedback as $\mathcal{L}_t:=\{l_{{s}|t}(\vec{a}_{{s}|t}):s+d_{{s}}=t,t\geq s\}\forall t\in\{1,...,T\}$. It is to be noted that the learner can observe the set of losses in $\mathcal{L}_t$ at time $t$ but does not know the exact time $s$ when those losses were actually chosen. Also, it is to be noted that we assume all losses are received at time $T$ despite the presence of delay.  If no loss is observed at time $t$, then $\mathcal{L}_t$ is an empty set. Traditionally, the learner is meant to select the probability vector $\{\vec{p}_s\}^{t=T}_{s=1}$ to minimize regret. However, the information to decide the $K\times 1$ probability vector, $\vec{p}_s$, may not be available due to the presence of delay. For instance, if the losses of an action selected at a given time is yet to be received, then it is difficult to update the probability vector that depends on those losses for the next round. However, the learner can utilize the available information collected in the set $\mathcal{L}_{1:t-1}=\cup^
{t-1}_{j=1}{\mathcal{L}_j}$ to compute $\vec{p}_t$ and not $\Vec{p}_s$. This is because the learner does not know $s$ when it receives $l_{s|t}(\Vec{a}_{s|t})$, but it does know $t$, so it can compute $\Vec{p}_t$ based on this knowledge.

The base algorithm EXP3.M cannot be applied in this setting with unknown delay since it is difficult to estimate the loss for the arms. This is explained in (\ref{eqn 2}) and (\ref{eqn 3}) below:

\begin{equation}
\quad\quad\quad
\hat{l}_{{s}|t}(i) = \frac{l_{{s}|t}(i)\mathbbm{I}(i
\in\vec{a}_{{s}|t})}{{p_t}{(i)}} \forall i\in \mathcal{A} .
\label{eqn 2}
\end{equation}
where $p_t(i)$ is the $i-th$ component of $\Vec{p}_t$,  and $l_{s|t}(i)$ is the true loss of  arm $i$ chosen at time $s$ and observed at time $t$ \color{black}.  For the semi-bandit setting discussed in this paper, the learner has access to the loss of only the arms s/he chooses contained in the loss vector $l_{s|t}(\Vec{a}_{s|t})$. This is the essence of the indicator function in (\ref{eqn 2}). The loss estimation is necessary because the learner does not know the value of the losses of other unchosen arms, yet it must update the probability over all arms based on these losses. The loss estimator  $\hat{l}_{{s}|t}{(i)}$ in (\ref{eqn 2}) is biased unlike in the classic setting because the action $\Vec{a}_{s|t}$ was originally drawn at time $s$ from the probability vector $\Vec{p}_s$ but the loss vector $l_{s|t}(\Vec{a}_{s|t})$ was received at time $t$, and the estimated loss is computed with the probability vector $\Vec{p}_t$. 

Hence, the  expectation of the loss estimator conditioned over the collection of past received losses and past played actions (represented with the filtration $\mathcal{F}_{t-1}:=\sigma(\mathcal{L}_{1:t},\Vec{a}_{s|1},...,\Vec{a}_{s|t-1})$) will not recover the true loss. For notation sake, we represent the conditional expectation $\mathbbm{E}_{\mathcal{F}_t/\mathcal{F}_{t-1}}[\hat{l}_{s|t}(i)|\mathcal{F}_{t-1}]$ simply as $\mathbbm{E}_{{a_s|t}\sim \Vec{p}_s}[\hat{l}_{s|t}(i)]$. Thus, the expectation of the loss estimate is given as: 
\begin{equation*}
    \quad\quad\quad
\mathbb{E}_{{\vec{a}_{s}|t}\sim \vec{p}_{s}}[\hat{l}_{{s}|t}(i)]=\mathbb{E}_{{\vec{a}_{s}|t}\sim \vec{p}_{s}}\Bigg[\frac{l_{{s}|t}(i)\mathbbm{I}(i
\in\vec{a}_{{s}|t})}{{p_t}{(i)}}\Bigg]
\end{equation*}

\begin{equation}
\begin{split}
  = \sum^K_{j=1}p_s(j)\frac{l_{s|t}(i)}{p_t(i)}\mathbb{I}(j=i: i\in\Vec{a}_{s|t})
  \\
  = \frac{l_{{s}|t}(i){p_{s}}(i)}{{p_t}(i)}\neq l_{{s}|t}(i) \forall i\in\vec{a}_{s_n|t}.
\label{eqn 3} 
\end{split}
\end{equation}

Here, $p_{s}(i)$ is the $i-th$ component of $\Vec{p}_s$. 

The regret for the delayed setting with multiple play for action $\Vec{a}_{s|t}$, chosen from the probability vector $\Vec{p}_t\in\Delta_K$, over all arms, is given as
\begin{equation}
    Reg^D_T := \sum^T_{t=1}\mathbbm{E}[\Vec{p}^\mathbb{T}_t\Vec{l}_t]-\sum^T_{t=1}(\Vec{p}^*)^\mathbb{T}\Vec{l}_t,
\end{equation}
where D means delayed. The expectation is over the randomness of $\Vec{p}_t$.

\section{Proposed Algorithm}
A randomized adversarial MAB algorithm is introduced in this section, named Delayed EXP3.M (DEXP3.M), that can handle multiple play settings with unknown delays.  In order to obtain the updated probability vector $\Vec{p}_{t+1}$, the learner has to update $|\mathcal{L}_t|$ times since there are multiple rounds of feedback contained in $\mathcal{L}_t$. There is no need for the time index $s$ in (\ref{eqn 2}) to be known for implementation of this algorithm. To upper bound the bias in (\ref{eqn 3}), there must be an upper bound on $\frac{p_{s}(i)}{p_{t}(i)}$ which results to a lower bound on $p_t{(i)}$.

The proposed DEXP3.M algorithm runs  \textbf{Depround} algorithm \cite{gandhi2006dependent}, as a subroutine for selecting $k$ arms out of $K$ total arms each time with linear time and space complexity, as it was done for the EXP3.M algorithm in \cite{uchiya2010algorithms}. The  Depround algorithm was used in \cite{uchiya2010algorithms}, to remove exponential complexity associated with multiple play setting in adversarial bandit. The input to the Depround algorithm had the sum of the probability distribution over $K$ arms equal to $k$, i.e., $\sum^K_{i=1}p_t(i)=k$.  This led to the problem of the probability of each arm exceeding one in the EXP3.M algorithm. This problem was addressed by cutting off weights of arms exceeding a certain threshold. This way, the probability of each arm, which depends on the weights of the arm, does not exceed one. The Depround algorithm updates $(p_t(1),...p_t(K))$ probabilistically until each probability $p_t(i)$, is either $1$ or $0$, while still ensuring that the sum of the probabilities is $k$. 

The DEXP3.M algorithm is different from the EXP3.M algorithm because it uses the probability simplex $\Delta_K$, whose total probability over $K$ arms is 1 and not $k$. This automatically removes the problem found in EXP3.M where an arm can have a probability that exceeds 1, and must have its weight cut off. For the DEXP3.M algorithm, no arm has a probability that exceeds one. However, the problem of using the Depround algorithm surfaces because the input to the Depround algorithm requires that the sum of the total probability be $k$. This problem can be overcome by scaling the probabilities over all arms by the factor $k$ at the point of execution of the Depround algorithm . There is no extra complexity incurred by this scaling, as the Depround algorithm runs at a linear complexity of $O(K)$. More so, there is no extra regret incurred since the analysis of the regret uses only the output of the Depround algorithm and excludes the operation of the Depround algorithm. The total probability is 1 outside the Depround algorithm and $k$ only inside the Depround algorithm. The proposed DEXP3.M algorithm also performs differently from the base EXP3.M algorithm by introducing a $K\times1$ weight vector $\vec{\Tilde{w}}_t$ to evaluate the performance of each arm historically as illustrated in (5) to (7). 

There are some similarities between DEXP3 algorithm in \cite{li2019bandit}, for single play adversarial bandit with delay, and the proposed DEXP3.M algorithm. The regret analysis for both algorithms uses the probability simplex, hence the regret definition of DEXP3 in  is the same as for DEXP3.M but with a different intuitive meaning, as shown in (4). An upper bound on $\frac{p_{s}(i)}{p_t(i)}$ is required for the analysis of both algorithms. There are also striking differences between both algorithms. First, DEXP3 algorithm does not use the Depround algorithm. Also,  the probability of each arm in DEXP3 algorithm is the normalized weight corresponding to that arm, but in DEXP3.M algorithm, the probability of each arm  is a trade-off between exploration with parameter $\gamma$ and exploitation with parameter $1-\gamma$. Hence, the learner either sticks to the arm that has given a good reward in the past or explore for new arm uniformly at random.  

For the implementation of the algorithm, we can assign numbers to all feedback losses received in $\mathcal{L}_{t}$, i.e., $\mathcal{L}_t :=\{l_{s_n|t}(\Vec{a}_{s_n|t}): \forall s_n=t-d_{s_n}\}$, $n=1,...,|\mathcal{L}_t|$. At time $t$, there are $|\mathcal{L}_t|$ feedback losses for the implementation of DEXP3.M algorithm. To update $\Vec{p}_{t+1}$ from $\Vec{p}_t$, the learner must update $|\mathcal{L}_t|$ times. The index of the inner loop update runs from $n=1$ to $n=|\mathcal{L}_t|$. The learner therefore updates $\vec{\Tilde{w}}_t$ starting from $\Vec{p}^{n-1=0}_{t}:=\Vec{p}_t$ to $\Vec{p}_t^{n =|\mathcal{L}_t|}:=\Vec{p}_{t+1}$ for each loss vector $l_{{s_n}|t}(\vec{a}_{{s_n}|t})\in \mathcal{L}_t$ using the estimated loss $\Vec{\hat{l}}_{{s_n}|t}$ over all arms,  as shown in (\ref{eqn 5})  below:\color{black}
\begin{equation}
\tilde{w}^n_t{(i)}=p^{n-1}_t{(i)}\exp(-\frac{k\gamma}{K}\min\{\delta_1,\hat{l}_{{s_n}|t}(i)\}), \forall i\in \mathcal{A};\delta_1\geq 0,
\label{eqn 5}
\end{equation}
where $\tilde{w}^n_t(i)$ is a component of $\tilde{\Vec{w}}_t$, $\gamma$ is the exploration parameter and $\delta_1$ is the upper bound on $\hat{l}_{{s_n}|t}(i)$ to control its bias. The parameter $\delta_1$ should be chosen to ensure that the probability of $\hat{l}_{{s_n}|t}(i)$ greater than $\delta_1$ tends to zero. The learner then finds the vector $\Vec{w}^n_t$ by a trimmed normalization as follows:
\begin{equation}
\quad\quad\quad
w^n_t{(i)}=\max\Bigg\{\frac{\Tilde{w}^n_t{(i)}}{\sum^K_{j=1}\Tilde{w}^n_t{(j)}},\frac{\delta_2}{K}\Bigg\},\forall i\in \mathcal{A}; \delta_2\geq 0.
\label{eqn 6}
\end{equation}
The update in (\ref{eqn 6}) ensures that $w^n_t{(i)}$ is lower bounded by $\frac{\delta_2}{K}$. The learner finally computes $\Vec{p}^n_t$ as:
\begin{equation}
    \quad\quad\quad
p^n_t{(i)}=(1-\gamma)\frac{w^n_t{(i)}}{\sum^K_{j=1}w^n_t{(j)}}+\frac{\gamma}{K},\forall i\in \mathcal{A}.
\label{eqn 7}
\end{equation}
When all elements of $\mathcal{L}_t$ has been used for the update, the learner finds $\vec{p}_{t+1}$ using:
\begin{equation}
\Vec{p}_{t+1}=\vec{p}^{|\mathcal{L}_t|}_{t}.
\label{eqn 8}
\end{equation}
However, if $\mathcal{L}_t=\emptyset$, the learner uses the previous distribution i.e., $\Vec{p}_{t+1}=\Vec{p}_t$ for choosing an arm. The DEXP3.M algorithm is summarized in Table 1 below.
\begin{table*}[h!]
\centering
    \begin{tabular}{l}
    \hline
DEXP3.M (The extended version of EXP3.M for bandit problems with unknown delays)\\
\hline
\textbf{Parameters}: $\gamma\in(0,1]$\\
\textbf{Initialization}: $p_1{(i)}=\frac{1}{K}$ $\forall i\in \mathcal{A}$\\
\textbf{For} $t = 1,...,T$\\
\quad Step 1: a. Scale the probabilities $(p_t(1),...,p_t(K))$ by a factor of $k$.\\
\quad\quad\quad\quad b. Use the scaled probabilities as input and select $k\leq K$ arms using \textbf{Depround} algorithm \cite{gandhi2006dependent}.\\
\quad\quad\quad\quad c. Return the original probabilities.\\
\quad Step 2: Observe feedback losses collected in $\mathcal{L}_t$ .\\
\quad( $\mathcal{L}_t=\{l_{{s_n}|t}(\vec{a}_{{s_n}|t}):s_n+d_{{s_n}}=t; \}),n=1,...,|\mathcal{L}_t|$.\\
\quad \textbf{If}  $\mathcal{L}_t$ $\neq \emptyset$; then\\
\quad\quad \textbf{For} $n=1,...,|\mathcal{L}_t|$\\

\quad\quad\quad Step i: Estimate loss vector $\hat{\vec{l}}_{{s_n}|t}$ via (2) $\Bigg( \hat{l}_{{s_n}|t}(i)= \frac{l_{{s_n}|t}(i)\mathbbm{I}(i
\in\vec{a}_{{s_n}|t})}{{p_t}{(i)}} \forall i\in \mathcal{A} \Bigg).$\\
\quad\quad\quad Step ii: Update $\vec{p}^n_t$ via (5) to (7).\\
\quad\quad\quad Step iii: Obtain $\Vec{p}_{t+1}= \vec{p}_t^{|\mathcal{L}_t|}$ via (8).\\
\quad\quad \textbf{End}\\
\quad \textbf{Else}\\
\quad\quad Step iv: $\vec{p}_{t+1}= \vec{p}_t$.\\

\quad \textbf{End}\\
\quad Step 3: Return to Step 1 and repeat for all $t$.\\
\textbf{End}\\
\hline
    \end{tabular}
    \caption{Pseudocode for algorithm DEXP3.M}
    \label{tab:my_label}
\end{table*}
\subsection{Mapping from real to virtual slot}

\begin{figure}
    \centering
    \includegraphics[width=4in]{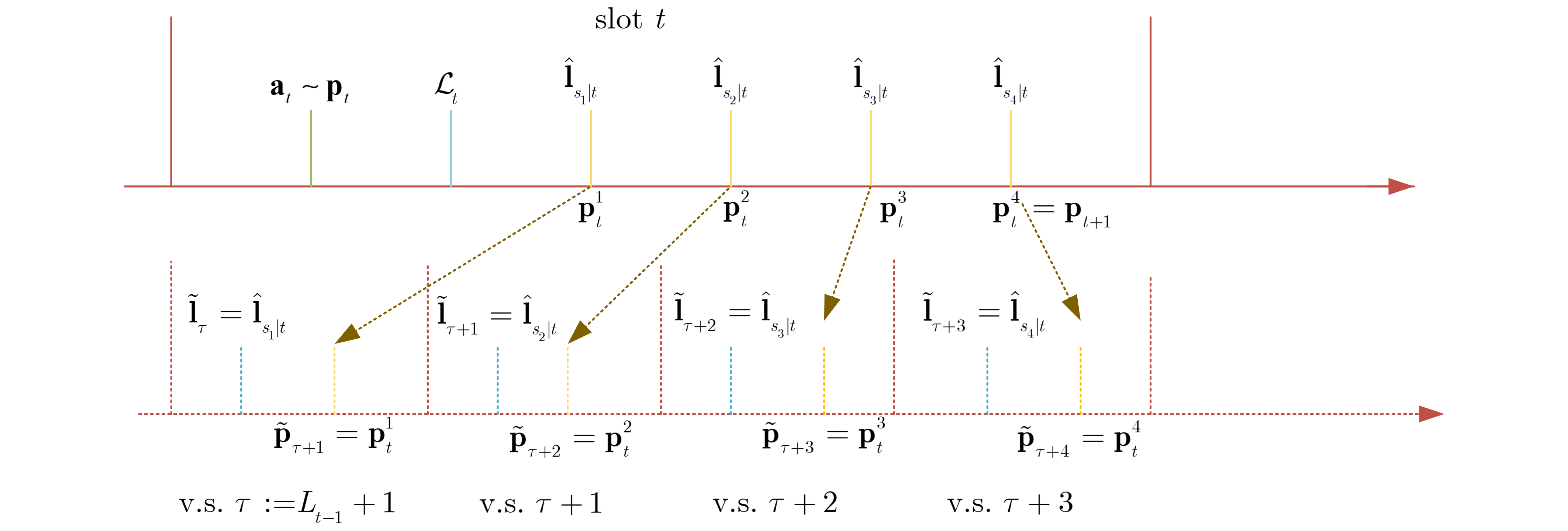}
    \caption{Mapping from real slots (solid line) to virtual slots (dotted line). The feedback $\mathcal{L}_t=\{l_{{s_1}|t}(\vec{a}_{{s_1}|t}),l_{{s_2}|t}(\vec{a}_{{s_2}|t}),l_{{s_3}|t}(\vec{a}_{{s_3}|t}),l_{{s_4}|t}(\vec{a}_{{s_4}|t})\}$ was received at time $t$.'v.s'. means virtual slot. Note that the losses may actually have been received without order and $t\geq5$.}
    \label{Drawing11}
\end{figure}

This paper overcomes the challenge of recursion between $\Vec{p}_t$ and $\Vec{p}_{t+1}$ by the introduction of         \say{virtual slots}. There are $T$ virtual timeslots over the real time horizon and the $\tau-th$ virtual slot is associated with the $\tau-th$ loss value fed back. The feedback received at the end of slot $t$ is $\mathcal{L}_t$ and the overall feedback sum received at the end of slot $t-1$ is $L_{t-1}:=\sum^{t-1}_{v=1}|\mathcal{L}_v|$. The virtual slot $\tau$ corresponding to the first feedback value received at slot $t$ can thus be written as $\tau=L_{t-1}+1$. When multiple rounds of feedback are received over a real timeslot $t$ , DEXP3.M updates $\vec{p}_t$ recursively $|\mathcal{L}_t|$ times to obtain $\vec{p}_{t+1}$ using (\ref{eqn 5}) to (\ref{eqn 8}). With the notion of virtual slot in mind, these $|\mathcal{L}_t|$ updates are done over $|\mathcal{L}_t|$ consecutive virtual slots. For instance, $\hat{\vec{l}}_{{s_1|t}}$ is used to update from $\vec{p}^0_t$ to $\Vec{p}^1_t$ and the update is mapped to virtual slot $\tau=L_{t-1}+1$. Hence, in the virtual slot,   $\Tilde{\vec{l}}_{\tau}=\hat{\Vec{l}}_{{s_1}|t}$ is used to obtain $\tilde{\Vec{p}}_{\tau+1}=\vec{p}^1_t$ from $\tilde{\vec{p}}_\tau = \vec{p}^0_t$. Similarly, $\vec{p}^2_t$ is obtained by using $\hat{\vec{l}}_{{s_2}|t}$ and the virtual slot yields $\Tilde{\vec{p}}_{\tau+2}=\vec{p}^2_t$ using $\Tilde{\vec{l}}_{\tau+1}=\hat{\vec{l}}_{{s_2}|t}$. This means that at real timeslot $t$, for $n=1,...,|\mathcal{L}_t|$, each update from $\vec{p}^{n-1}_t$ to $\vec{p}^n_t$ using $\hat{\vec{l}}_{{s_n}|t}$ is mapped to an update at the virtual slot $\tau+n-1$, where $\Tilde{\Vec{l}}_{\tau+n-1}=\hat{\vec{l}}_{{s_n}|t}$ is used to obtain $\Tilde{\Vec{p}}_{\tau+n}=\vec{p}^n_t$ from $\Tilde{\vec{p}}_{\tau+n-1}=\vec{p}^{n-1}_t$. From the real-to-virtual slot mapping, it can be seen that $\Tilde{\vec{p}}_{\tau+|\mathcal{L}_t|}=\vec{p}_{t+1}$.  This is illustrated in Fig. 1 for the situation where the cardinality of $\mathcal{L}_t=4$. Since many feedback can be received at any time without any particular order, the learner may know the time interval when the actions producing the feedback were chosen, but he cannot match the actions to the time when the feedback was actually incurred. Analyzing the recursion between two consecutive $\Tilde{\vec{p}}_{\tau}$ and $\Tilde{\vec{p}}_{\tau+1}$ will be paramount for the regret analysis. 

Updating in the virtual slot $\tau$ is same as in (\ref{eqn 5}), (\ref{eqn 6}) and (\ref{eqn 7})

\begin{equation}
\tilde{w}_{\tau + 1}(i)=\tilde{p}_{\tau}(i)\exp(-\frac{k\gamma}{K}\min\{\delta_1,\tilde{l}_\tau(i)\}),\forall i\in \mathcal{A};\delta_1\geq 0  
\label{eqn 9}
\end{equation}

\begin{equation}
w_{\tau+1}(i)=\max\Bigg\{\frac{\tilde{w}_{\tau+1}(i)}{\sum^K_{j=1}\tilde{w}_{\tau+1}(j)},\frac{\delta_2}{K}\Bigg\},\forall i\in \mathcal{A}; \delta_2\geq0 
\label{eqn 10}
\end{equation}

\begin{equation}
\tilde{p}_{\tau+1}(i)=(1-\gamma)\frac{w_{\tau+1}(i)}{\sum^K_{j=1}{w_{\tau+1}(j)}}+\frac{\gamma}{K},\quad \forall i\in \mathcal{A} 
\label{eqn 11}
\end{equation}

\begin{equation}
\sum^K_{j=1}\tilde{w}_{\tau}(j)\leq\sum^K_{j=1}\tilde{p}_{\tau-1}(j){= 1}. 
\label{eqn 12}
\end{equation}

And $\sum^K_{i=1}w_\tau(i)$ is upper and lower bounded by

\begin{equation}
\sum^K_{i=1}w_{\tau}(i)\geq \sum^K_{i=1}\frac{\tilde{w}_{\tau}(i)}{\sum^K_{j=1}\tilde{w}_{\tau}(j)}=1
\label{eqn 13}
\end{equation}

\begin{equation}
\sum^K_{i=1}w_{\tau}(i)\leq \sum^K_{i=1}\frac{\tilde{w}_\tau(i)}{\sum^K_{j=1}\tilde{w}_\tau(j)} +\delta_2 = 1+\delta_2,  
\label{eqn 14}
\end{equation}

\begin{equation}
\sum^K_{i=1}\tilde{p}_{\tau}(i)  \leq^{(a)} \sum^K_{i=1}w_\tau(i)
 \label{eqn 15}
\end{equation}

(a) follows from (\ref{eqn 13}). 

\begin{equation}
\begin{split}
(1-\gamma)\frac{\delta_2}{K(1+\delta_2)}+\frac{\gamma}{K}\leq^{(b)}(1-\gamma)\frac{w_\tau(i)}{1+\delta_2}+\frac{\gamma}{K}\leq^{(c)}\\
\tilde{p}_\tau(i)\leq w_\tau(i)
\label{eqn 16}
\end{split}
\end{equation}

(b) and (c) is deduced from (\ref{eqn 10}), (\ref{eqn 11}) and (\ref{eqn 14}).

\textbf{Assumption 1}: The maximum loss is upper bounded as follows:
\begin{equation}
 \max_{t,i}l_{t}(i)\leq 1.   
\end{equation}

\textbf{Assumption 2}: The delay $d_{t}$ is upper bounded as follows:
\begin{equation}
    \max_{t}d_t\leq \Bar{d}.
\end{equation}

\textbf{Lemma 1}: In consecutive virtual slot $\tau -1$ and $\tau$, the following inequalities hold for any $i$
\begin{equation*}
 \tilde{p}_{\tau-1}(i)-\tilde{p}_\tau(i)\leq \tilde{p}_{\tau-1}(i)\Bigg[\frac{\gamma+\frac{k\gamma}{K}\min\{\delta_1,\tilde{l}_{\tau-1}(i)\}+\delta_2}{1+\delta_2}\Bigg]  
\end{equation*}

and
\begin{equation}
\frac{\Tilde{p}_{\tau-1}(i)}{\Tilde{p}_{\tau}(i)}\leq\frac{1}{1-\gamma-\frac{k\gamma\delta_1}{K}-\delta_2}      
\end{equation}
if the parameters are chosen properly such that $1-\gamma-\frac{k\gamma\delta_1}{K}-\delta_2\geq0$. 

\textbf{Proof}:
First we have
\begin{equation}
\begin{split}
\tilde{p}_\tau(i)\geq^{(a)}(1-\gamma)\frac{w_\tau(i)}{1+\delta_2}+\frac{\gamma}{K}\geq^{(b)}\\
(1-\gamma)\frac{\tilde{w}_\tau(i)}{\sum^K_{j=1}\tilde{w}_\tau(j)(1+\delta_2)}+\frac{\gamma}{K}    
\end{split}
\end{equation}

\begin{equation}
 \geq^{(c)}(1-\gamma)\frac{\tilde{w}_\tau(i)}{1+\delta_2}+\frac{\gamma}{K}
\end{equation}

\begin{equation*}
\quad {\geq} \frac{(1-\gamma)[\tilde{p}_{\tau-1}(i)\exp(-\frac{k\gamma}{K}\min\{\delta_1,\tilde{l}_{\tau-1}(i)\})]}{1+\delta_2}+\frac{\gamma}{K}   
\end{equation*}

\begin{equation*}
\begin{split}
-\tilde{p}_{\tau}(i)\leq^{(d)}
\frac{(1-\gamma)[\tilde{p}_{\tau-1}(i)(-1+\frac{k\gamma}{K}\min\{\delta_1,\tilde{l}_{\tau-1}(i)\})]}{1+\delta_2} -\\
\frac{\gamma}{K}
\end{split}
\end{equation*}
upper bounding and adding $\tilde{p}_{\tau-1}(i) $ on both side of the equation,
\begin{equation}
\begin{split}
\tilde{p}_{\tau-1}(i)-\tilde{p}_\tau(i)\leq\\
\frac{(1-\gamma)[\Tilde{p}_{\tau-1}(i)(-1+\frac{k\gamma}{K}\min\{\delta_1,\tilde{l}_{\tau-1}(i)\})]}{1+\delta_2} + \\
\tilde{p}_{\tau-1}(i) 
\end{split}
\end{equation}

\begin{equation*}
\begin{split}
 \tilde{p}_{\tau-1}(i)-\tilde{p}_\tau(i)\leq\\ \tilde{p}_{\tau-1}(i)\Bigg[\frac{(1-\gamma)(-1+\frac{k\gamma}{K}\min\{\delta_1,\tilde{l}_{\tau-1}(i)\})}{1+\delta_2}+1\bigg] 
 \end{split}
\end{equation*},


\begin{equation}
\leq\tilde{p}_{\tau-1}(i)\Bigg[\frac{\gamma+\frac{k\gamma}{K}\min\{\delta_1,\tilde{l}_{\tau-1}(i)\}+\delta_2}{1+\delta_2}\Bigg]
\label{eqn 23}  
\end{equation}
which is the first inequality of Lemma 1. To obtain the second inequality of Lemma 1, (\ref{eqn 23}) is further upper bounded and the ratio of two consecutive probabilities is obtained as follows\\

\begin{equation*}
\tilde{p}_{\tau-1}(i)-\tilde{p}_\tau(i)\leq \Tilde{p}_{\tau-1}(i)[\gamma+\frac{k\gamma\delta_1}{K} +\delta_2]
\end{equation*}
\begin{equation}
\frac{\Tilde{p}_{\tau-1}(i)}{\Tilde{p}_{\tau}(i)}\leq\frac{1}{1-\gamma-\frac{k\gamma\delta_1}{K}-\delta_2}    
\end{equation}

In (a),(b) and (c) we use (\ref{eqn 16}), (\ref{eqn 10}) and (\ref{eqn 12}) respectively while in (d), we first use $\exp(-x)\geq 1-x$ and then multiply the equation by $-1$.

\textbf{Lemma 2}: The following inequality holds for any $\tau$ and any $i$
\begin{equation}
\begin{split}
\tilde{p}_{\tau}(i)-\tilde{p}_{\tau-1}(i)\\
\leq \tilde{p}_{\tau}(i)\Bigg[1-I_\tau(i)\sum^K_{j=1}\tilde{p}_{\tau-1}(j)(1-\frac{k\gamma}{K}\min\{\delta_1,\tilde{l}_{\tau-1}(j)\})\Bigg] 
\label{eqn 25}
\end{split}
\end{equation}

where $I_\tau(i)=\mathbbm{I}(w_\tau(i)>\frac{\delta_2}{K})$.

\textbf{Proof}:
An intermediate result to be shown for the proof is given as
\begin{equation}
 \tilde{w}_\tau(i)\geq \tilde{p}_\tau(i)I_\tau(i)\sum^K_{j=1}\tilde{w}_{\tau}(j)   
 \label{eqn 26}
\end{equation}

When $I_\tau(i)=0$, the inequality holds by inspection. When $I_\tau(i)=1$, $w_\tau(i)=\tilde{w}_\tau(i)/\sum^K_{j=1}\tilde{w}_\tau(j)$. From (\ref{eqn 16}),  $\tilde{p}_\tau(i)\leq w_\tau(i)=
\tilde{w}_\tau(i)/\sum^K_{j=1}\tilde{w}_\tau(j)$ and (\ref{eqn 26}) is established. Using this intermediate result,

\begin{equation*}
\begin{split}
 \tilde{p}_\tau(i)-\tilde{p}_{\tau-1}(i)\leq^{(a)} \tilde{p}_\tau(i)-\tilde{w}_\tau(i)\\
 \leq^{(b)} \tilde{p}_\tau(i)-\tilde{p}_\tau(i)I_\tau(i)\sum^K_{j=1}\tilde{w}_\tau(j)   
 \end{split}
\end{equation*}
\begin{equation*}
 =\tilde{p}_\tau(i)\Bigg[1-I_\tau(i)\sum^K_{j=1}\tilde{p}_{\tau-1}(j)\exp\Bigg(-\frac{k\gamma}{K}\min\{\delta_1,\tilde{l}_{\tau-1}(j)\}\Bigg)\Bigg]   
\end{equation*}

\begin{equation}
\leq^{(c)} \tilde{p}_\tau(i)\Bigg[1-I_\tau(i)\sum^K_{j=1}\tilde{p}_{\tau-1}(j)\Bigg(1-\frac{k\gamma}{K}\min\{\delta_1,\tilde{l}_{\tau-1}(j)\}\Bigg)\Bigg]  
\label{eqn 27}
\end{equation}

where in (a) and (b), (\ref{eqn 9}) and (\ref{eqn 26}) is used. In (c), $\exp\{-x\}\geq1-x$ is used.

\textbf{Lemma 3}: The following inequality holds
\begin{equation}
\frac{\tilde{p}_\tau(i)}{\tilde{p}_{\tau-1}(i)}\leq \max\Bigg\{\frac{\delta_2(1+\delta_2)}{\gamma+\delta_2},\frac{1}{1-\frac{k\gamma\delta_1}{K}}\Bigg\}   
\end{equation}.

\textbf{Proof}: The proof of Lemma 3 is based on the result of Lemma 2. Consider first the case where $I_\tau(i)=0$. Lemma 2 becomes $\Tilde{p}_\tau(i)-\tilde{p}_{\tau-1}(i)\leq \tilde{p}_\tau(i)$ which is an obvious inconsequential fact. However, when $I_\tau(i)=0$, then the upper bound on $w_\tau(i)=\frac{\delta_2}{K}$ based on the indicator function in Lemma 2. By applying (\ref{eqn 16}), $\tilde{p}_\tau(i)\leq w_\tau(i)=\frac{\delta_2}{K}$. Hence, 

\begin{equation*}
 \frac{\tilde{p}_\tau(i)}{\tilde{p}_{\tau-1}(i)}\leq \frac{\delta_2}{K}\frac{1}{\tilde{p}_{\tau-1}(i)},
\end{equation*}
inserting the lower bound of $\tilde{p}_{\tau-1}(i)$ from (\ref{eqn 16}) results into

 

\begin{equation}
 =\frac{\delta_2(1+\delta_2)}{\gamma +\delta_2}.
 \label{eqn 29}  
\end{equation}

Now, considering when $I_\tau(i)=1$, Lemma 2 becomes

\begin{equation*}
\begin{split}
  \tilde{p}_\tau(i)-\tilde{p}_{\tau-1}(i)\leq \tilde{p}_\tau(i)\\
  \Bigg[1-\sum^K_{j=1}\tilde{p}_{\tau-1}(j)\Bigg(1-\frac{k\gamma}{K}\min\{\delta_1,\tilde{l}_{\tau-1}(j)\}\Bigg)\Bigg]
  \end{split}
\end{equation*}

\begin{equation*}
 = \tilde{p}_\tau(i)\Bigg[1-\sum^K_{j=1}\tilde{p}_{\tau-1}(j)+\frac{k\gamma}{K}\sum^K_{j=1}\tilde{p}_{\tau-1}(j)\min\{\delta_1,\tilde{l}_{\tau-1}(j)\}\Bigg]  
\end{equation*}

\begin{equation*}
=^{(a)}\frac{k\gamma}{K}\tilde{p}_\tau(i)\sum^K_{j=1}\tilde{p}_{\tau-1}(j)\min\{\delta_1,\tilde{l}_{\tau-1}(j)\}    
\end{equation*}

in (a), (\ref{eqn 12}) is used.
\color{black}
\begin{equation*}
 \leq \frac{k\gamma}{K}\tilde{p}_\tau(i)\sum^K_{j=1}\tilde{p}_{\tau-1}(j)\delta_1   
\end{equation*}


\begin{equation*}
-\tilde{p}_{\tau-1}(i)\leq[\frac{k\gamma\delta_1}{K}-1]\tilde{p}_\tau(i)   
\end{equation*}

Hence,
\begin{equation}
 \frac{\tilde{p}_{\tau}(i)}{\tilde{p}_{\tau-1}(i)}\leq\frac{1}{1-\frac{k\gamma \delta_1}{K}}.
 \label{eqn 30}
\end{equation}

Combining (\ref{eqn 29}) and (\ref{eqn 30}) the proof is completed.

\textbf{Lemma 4}:
For a given sequence of loss $\{\tilde{\vec{l}}_\tau\}^T_{\tau=1}$, the following holds

\begin{equation}
\begin{split}
\sum^T_{\tau=1}(\tilde{\Vec{p}}_\tau-\Vec{p})\min\{\tilde{\Vec{l}}_\tau,\delta_1\cdot \Vec{1}\}
\leq\\
\frac{\ln{K}+T\ln({1+\delta_2)}}{(\frac{k\gamma}{K})}+\frac{k\gamma}{2K}\sum^T_{\tau=1}\sum^K_{i=1}\tilde{p}_\tau(i)[\tilde{l}_\tau(i)]^2 
\end{split}
\end{equation}
where $\Vec{1}$ is a $K\times 1$ vector of all ones, and $\Vec{p}\in{\bigtriangleup}_K$.

\textbf{Proof}:
Let $\tilde{\Vec{c}}_\tau = \min\{\tilde{\Vec{l}}_\tau,\delta_1\cdot\Vec{1}\}$. Hence, $\tilde{c}_\tau(i)=\min\{\tilde{l}_\tau(i),\delta_1\}$. Also, let $\tilde{W}_\tau=\sum^K_{i=1}\tilde{w}_\tau(i)$ and $W_\tau=\sum^K_{i=1}w_\tau(i)$.

\begin{equation*}
 \tilde{W}_{T+1}=\sum^K_{i=1}\tilde{w}_{T+1}(i)=\sum^K_{i=1}\tilde{p}_T(i)\exp(\frac{-k\gamma}{K}\tilde{c}_T(i))   
\end{equation*}

\begin{equation*}
 =\sum^K_{i=1}\Bigg[(1-\gamma)\frac{w_T(i)}{W_T}+\frac{\gamma}{K}\Bigg]\exp{\Bigg(\frac{-k\gamma}{K}\tilde{c}_T(i)\Bigg)}   
\end{equation*}

\begin{equation*}
 \geq \sum^K_{i=1}\Bigg[(1-\gamma)\frac{\tilde{w}_T(i)}{\tilde{W}_T {W_T}}  +\frac{\gamma}{K}\Bigg]\exp\Bigg(\frac{-k\gamma}{K}\tilde{c}_T(i)\Bigg) 
\end{equation*}

\begin{equation*}
\geq \sum^K_{i=1}\Bigg[(1-\gamma)\frac{\tilde{w}_T(i)}{\tilde{W}_T{W_T}}\Bigg]\exp\Bigg(\frac{-k\gamma}{K}\tilde{c}_T(i)\Bigg)    
\end{equation*}

\begin{equation*}
 =\sum^K_{i=1}\Bigg[(1-\gamma)\frac{\tilde{p}_{T-1}(i)\exp(\frac{-k\gamma}{K}\tilde{c}_{T-1}(i))}{\tilde{W}_T{W_T}}\Bigg]\exp\Bigg(\frac{-k\gamma}{K}\tilde{c}_T(i)\Bigg) \end{equation*}

\begin{equation*}
 =\sum^K_{i=1}(1-\gamma)\tilde{p}_{T-1}(i)\frac{\exp{[\frac{-k\gamma}{K}\tilde{c}_T(i)-\frac{k\gamma}{K}\tilde{c}_{T-1}(i)]}}{\tilde{W}_T{W_T}}   
\end{equation*}

\begin{equation*}
\begin{split}
= \sum^K_{i=1}(1-\gamma)\Bigg[(1-\gamma)\frac{w_{T-1}(i)}{W_{T-1}}+\frac{\gamma}{K}\bigg]\times\\
\frac{\exp{[\frac{-k\gamma}{K}\tilde{c}_T(i)-\frac{k\gamma}{K}\tilde{c}_{T-1}(i)]}}{\tilde{W}_T{W_T}} \end{split}
\end{equation*}

\begin{equation*}
\begin{split}
\geq \sum^K_{i=1}(1-\gamma)\Bigg[(1-\gamma)\frac{\tilde{w}_{T-1}(i)}{\tilde{W}_{T-1}W_{T-1}}\Bigg]\times \\
\frac{\exp{[\frac{-k\gamma}{K}\tilde{c}_T(i)-\frac{k\gamma}{K}\tilde{c}_{T-1}(i)]}}{\tilde{W}_T{W_T}} \geq\dots \end{split}
\end{equation*}

\begin{equation}
\geq (1-\gamma)^T\frac{\sum^K_{i=1}\tilde{w}_1(i)\exp\Bigg[-\frac{k\gamma}{K}\sum^T_{\tau=1}\tilde{c}_\tau(i)\bigg]}{\prod^T_{\tau=1}(W_\tau{\tilde{W}_\tau})} 
\label{eqn 32}
\end{equation}
For any probability distribution $\vec{p}\in\bigtriangleup_K$ with $\tilde{w}_1(i)=1$ and $\tilde{W}_1=K$, (\ref{eqn 32}) becomes

\begin{equation*}
\begin{split}
(1-\gamma)^T\sum^K_{i=1}p(i)\exp{\Bigg[-\frac{k\gamma}{K}\sum^T_{\tau=1}\tilde{c}_\tau(i)\Bigg]}\leq\\ (1-\gamma)^T\sum^K_{i=1}\exp{\Bigg[-\frac{k\gamma}{K}\sum^T_{\tau=1}\tilde{c}_\tau(i)\Bigg]}
\end{split}
\end{equation*}

\begin{equation}
 \leq \tilde{W}_1\prod^T_{\tau=1}(W_\tau{\tilde{W}_{\tau+1}})\leq^{(a)}K(1+\delta_2)^T\prod^{T}_{\tau=1}\tilde{W}_{\tau+1}   
 \label{eqn 33}
\end{equation}

in (a) we use $W_\tau\leq 1+\delta_2$ from (\ref{eqn 14}). Using Jensen's inequality
\begin{equation}
\begin{split}
(1-\gamma)^T\sum^K_{i=1}p(i)\exp{\Bigg[-\frac{k\gamma}{K}\sum^T_{\tau=1}\tilde{c}_\tau(i)\Bigg]}  \geq\\ (1-\gamma)^T\exp\Bigg[-\frac{k\gamma}{K}\sum^K_{i=1}\sum^T_{\tau=1}p(i)\tilde{c}_\tau(i)\Bigg]  
\label{eqn 34}
\end{split}
\end{equation}

substituting (\ref{eqn 34}) into (\ref{eqn 33})
\begin{equation}
\begin{split}
 (1-\gamma)^T\exp\Bigg[-\frac{k\gamma}{K}\sum^K_{i=1}\sum^T_{\tau=1}p(i)\tilde{c}_\tau(i)\Bigg]  \leq\\ K(1+\delta_2)^T\prod^T_{\tau=1}\tilde{W}_{\tau+1}. 
 \end{split}
 \label{eqn 35}
\end{equation}

On the other hand,
\begin{equation*}
\tilde{W}_{\tau+1}=\sum^K_{i=1}\tilde{w}_{\tau+1}(i) = \sum^K_{i=1}\tilde{p}_{\tau}(i)\exp{(-\frac{k\gamma}{K}\tilde{c}_{\tau}(i))}    
\end{equation*}

\begin{equation*}
 \leq^{(b)} \sum^K_{i=1}\tilde{p}_{\tau}(i)\bigg(1-\frac{k\gamma}{K}\tilde{c}_{\tau}(i)+\frac{k^2\gamma^2}{2K^2}[\tilde{c}_{\tau}(i)]^2\Bigg)   
\end{equation*}

where (b) follows from $\exp(-x)\leq 1-x+\frac{x^2}{2}$, $\forall x\geq 0,$

 Taking the log of both sides
\begin{equation*}
\ln{\tilde{W}}_{\tau+1} \leq \ln{\Bigg(1-\frac{k\gamma}{K}\sum^K_{i=1}\tilde{p}_{\tau}(i)\tilde{c}_{\tau}(i) + \frac{k^2\gamma^2}{2K^2}\sum^K_{i=1}\tilde{p}_{\tau}(i)[\tilde{c}_{\tau}(i)]^2\Bigg)}    
\end{equation*}

\begin{equation}
\leq^{(d)}-\frac{k\gamma}{K}\sum^K_{i=1}\tilde{p}_{\tau}(i)\tilde{c}_{\tau}(i)  + \frac{k^2\gamma^2}{2K^2}\sum^K_{i=1}\tilde{p}_{\tau}(i)[\tilde{c}_{\tau}(i)]^2    
\label{eqn 36}
\end{equation}
(d) follows from $\ln(1+x)\leq x .$

Taking the log of (\ref{eqn 35}) and substituting (\ref{eqn 36}) into (\ref{eqn 35})
\begin{equation}
\begin{split}
T\ln(1-\gamma) - \frac{k\gamma}{K}\sum^K_{i=1}\sum^T_{\tau=1}p(i)\tilde{c}_\tau(i)\leq\\ \ln{K}+T\ln{(1+\delta_2)}-\frac{k\gamma}{K}\sum^T_{\tau=1}\sum^K_{i=1}\tilde{p}_\tau(i)\tilde{c}_\tau(i)+ \\
  \frac{k^2\gamma^2}{2K^2}\sum^T_{\tau=1}\sum^K_{i=1}\tilde{p}_\tau(i)[\tilde{c}_\tau(i)]^2. 
\end{split}
\label{eqn 37}
\end{equation}

Rearranging (\ref{eqn 37}) and writing in vector form

\begin{equation*}
\begin{split}
\frac{k\gamma}{K}\sum^T_{\tau=1}(\Vec{\tilde{p}}_\tau - \Vec{p})^T\tilde{\Vec{c}}_\tau \leq \ln{K}+T\ln{(1+\delta_2)}-T\ln(1-\gamma)+\\
\frac{k^2\gamma^2}{2K^2}\sum^T_{\tau=1}\sum^K_{i=1}\tilde{p}_\tau(i)[\tilde{c}_\tau(i)]^2   \end{split} 
\end{equation*}
By upper bounding to remove the negative term and dividing by $\frac{k\gamma}{K}$
\begin{equation*}
\begin{split}
\sum^T_{\tau=1}(\tilde{\vec{p}}_\tau-\Vec{p})^T\tilde{c}_\tau \leq^{(e)} \frac{\ln{K}+T\ln{(1+\delta_2)}}{(\frac{k\gamma}{K})}+\\
\frac{k\gamma}{2K}\sum^T_{\tau=1}\sum^K_{i=1}\tilde{p}_\tau(i)[\tilde{c}_\tau(i)]^2  \end{split}  
\end{equation*}

\begin{equation}
\leq \frac{\ln{K}+T\ln(1+\delta_2)}{(\frac{k\gamma}{K})}+ \frac{k\gamma}{2K}\sum^T_{\tau=1}\sum^K_{i=1}\tilde{p}_\tau(i)[\tilde{l}_\tau(i)]^2    
\end{equation}

\textbf{Lemma 5} (Lemma 6 in \cite{li2019bandit}):
Let $\tilde{s}_\tau \triangleq \tau-1-L_{t(\tau)-1}$, and let $t(\tau)$ denote the real slot when the real loss $ \Vec{l}_{t(\tau)}$ corresponding to $\tilde{\Vec{l}}_\tau$ was originally incurred, i.e., $\tilde{\Vec{l}}_\tau= \hat{\Vec{l}}_{t(\tau)|t(\tau)+d_{t(\tau)}}$. Then, the following holds : \\
(i) $\tilde{s}_\tau\geq 0, \forall \tau$; (ii) $\sum^T_{\tau=1}\tilde{s}_{\tau}=\sum^T_{t=1}d_t$; and (iii) if $\max_t{d_t}\leq \Bar{d}$, then $\tilde{s}_\tau\leq 2\Bar{d}, \forall \tau$.

\textbf{Proof}:
(i) Observe Figure 2 and Table 2 shown below.
Notice at virtual slot $\tau$, the observed loss at the real slot is $l_{t(\tau)}(\Vec{a}_{t(\tau)})$ and the corresponding $\tilde{s}_\tau = \tau - 1 - L_{t(\tau)-1}$. If there are $m$ feedback received from time $1$ to $t(\tau)-1$, then $L_{t(\tau)-1}=m$ and $0\leq m \leq t(\tau)-1$ (from the meaning of $L_{t(\tau)-1}$). At the start of $t_1 = t(\tau)$, there are thus $m$ received feedback. However, due to delay, the feedback  $l_{t(\tau)}(\Vec{a}_{t(\tau)})$, for the action chosen at $t_1 = t(\tau)$, is received at $t_2 = t(\tau)+d_{t(\tau)}\geq t_1$. Hence, at the start of $t_2$, there are at least $m$ observations. This means $\tau \geq m+1$ and $\tilde{s}_\tau \geq m+1-1-m = 0$.\\

\begin{table*}[t!]
\centering
    \begin{tabular}{c| c| c| c}
    \hline
Virtual Slot  & $\tau=1$ & $\tau=2$  & $\tau=3$\\
\hline
$t(\tau)$ & 2 & 3 & 1   \\
\hline
$L_{t(\tau)-1}$  & 0  & 1 & 0\\
\hline
$\tilde{s}_\tau$  & 0  & 0 & 2\\
\hline
    \end{tabular}
    \caption{The values of $t(\tau)$, $L_{t(\tau)-1}$ and $\tilde{s}_\tau$ in Figure 1}
    \label{tab:my_label}
\end{table*}
(ii) 
\begin{equation}
\begin{split}
 \sum^T_{\tau=1} \tilde{s}_\tau= \sum^T_{\tau=1}(\tau-1-L_{t(\tau)-1})\\
\\
\quad\quad\quad=^{(a)}\sum^T_{t=1}(t-1-L_{t-1}) =^{(b)}\sum^T_{t=1}d_t    
\end{split}
\end{equation}
where (a) is because $\{{t(\tau)}\}^T_{\tau=1}$ is a permutation of $\{1,...,T\}$; and (b) is from the definition of $L_{t-1}$.\\
(iii) Since the losses of slots $t\leq t(\tau)-1-\Bar{d}$ must have been received at the beginning of $t=t(\tau)$; therefore $L_{t(\tau)-1}\geq t(\tau)-1-\Bar{d}$, and correspondingly 
\begin{equation}
\tilde{s}_\tau = \tau-1-L_{t(\tau)-1}\leq \tau-1-t(\tau)+1+\Bar{d}\leq^{(c)}2\Bar{d} 
\label{eqn 40}
\end{equation}
where (c) follows that $l_{t(\tau)}(\vec{a}_{t(\tau)})$ is observed at the end of slot $t= t(\tau)+d_{t(\tau)}$, and $L_{t(\tau)+d_{t(\tau)}-1}$ is at most $t(\tau)+d_{t(\tau)}-1$.   Hence, $\tau \leq t(\tau)+d_{t(\tau)}$ which results into $\tau - t(\tau) \leq d_{t(\tau)} \leq \Bar{d}$.

\begin{figure}[h!]
\centering
\includegraphics[width=3in]{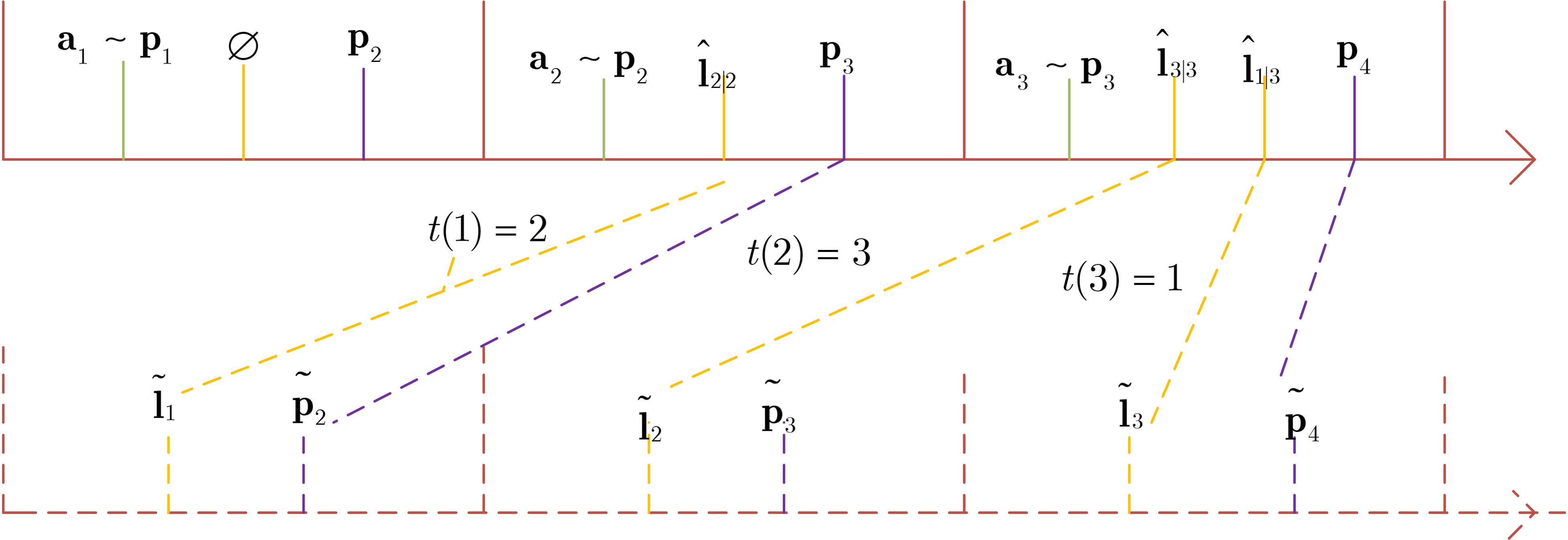}
\caption{Example of mapping from real slots (solid line) to virtual slots ( dotted line). The value of $t(\tau)$ is shown beside the corresponding yellow dotted arrow; $T = 3$ with delay $d_1=2$, $d_2 =0$ and $d_3=0$.}
\label{fig1}
\end{figure}

\textbf{Theorem}: If the total delay $D = \sum^T_{t=1}d_t$ and choosing $\delta_2 = \frac{1}{T+D}$, $\gamma = \sqrt{\frac{K(1+\ln{K})}{k^3\Bar{d}(T+D)}}$ and $\delta_1 = \frac{1}{2\gamma\bar{d}}+\frac{\delta_2}{\gamma}$, then the upper bound on regret of DEXP3.M is
\begin{equation}
    Reg^{D}_T = \mathcal{O}\sqrt{\bar{d}k(T+D)K(1+\ln K)}.
\end{equation}
\textbf{Proof}: The instantaneous regret is given by
\begin{equation*}
\Vec{p}^\mathbb{T}_t\Vec{l}_t -\Vec{p}^\mathbb{T}\Vec{l}_t = \sum^K_{i=1}p_t(i)l_t(i)-\sum^K_{i=1}p(i)l_t(i)
\end{equation*}

\begin{equation*}
\begin{split}
=^{(a)}\sum^K_{i=1}p_t(i)\mathbbm{E}_{\Vec{a}_t}\Bigg[\frac{l_t(i)\mathbbm{I}(i\in \Vec{a}_t)}{p_t(i)}\Bigg]  -\\
\sum^K_{i=1}p(i)\mathbbm{E}_{\Vec{a}_t}\Bigg[\frac{l_t(i)\mathbbm{I}(i\in \Vec{a}_t)}{p_t(i)}\Bigg] 
\end{split}
\end{equation*}

\begin{equation*}
= \sum^K_{i=1}(p_t(i)-p(i))\mathbbm{E}_{\Vec{a}_t}\Bigg[\frac{l_t(i)\mathbbm{I}(i\in\Vec{a}_t) }{p_{t+d_t}(i)}\frac{p_{t+d_t}(i)}{p_t(i)}\Bigg]   
\end{equation*}

\begin{equation*}
\leq \max_i \frac{p_{t+d_t}(i)}{p_t(i)}\sum^K_{i=1}(p_t(i)-p(i))\mathbbm{E}_{\vec{a}_t}\Bigg[\frac{l_t(i)\mathbbm{I}(i\in \vec{a}_t)}{p_{t+d_t}(i)}\Bigg]
\end{equation*}

\begin{equation}
 =^{(b)}\Bigg(\max_i \frac{p_{t+d_t}(i)}{p_t(i)}\Bigg)\mathbbm{E}_{\Vec{a}_t}\Bigg[\Vec{p}^\mathbb{T}_t \hat{\Vec{l}}_{t|t+d_t}-\Vec{p}^\mathbb{T}\hat{\Vec{l}}_{t|t+d_t}\Bigg]   
\end{equation}
where (a) is obtained from $\mathbbm{E}_{\Vec{a}_t}\Bigg[\frac{l_t(i)\mathbbm{I}(i\in\Vec{a}_t)}{p_t(i)}\Bigg]=l_t(i)$, and (b) is obtained from $\hat{l}_{t|t+d_t}(i)=\frac{l_t(i)\mathbbm{I}(i\in\Vec{a}_t)}{p_{t+d_t}(i)}.$
The overall regret over $T$ slots is given by
\begin{equation*}
\begin{split}
 Reg^D_T = \mathbbm{E}\Bigg[\sum^T_{t=1}\vec{p}^\mathbb{T}_t \Vec{l}_t - \Vec{p}^\mathbb{T}\Vec{l}_t\Bigg]\\ \leq \mathbbm{E}\Bigg[\sum^T_{t=1}\Bigg(\max_i \frac{p_{t+d_t}(i)}{p_t(i)}\Bigg)\mathbbm{E}_{\Vec{a}_t}\Bigg[\Vec{p}^\mathbb{T}_t\hat{\Vec{l}}_{t|t+d_t}-\Vec{p}^\mathbb{T}\hat{\Vec{l}}_{t|t+d_t}\Bigg]\Bigg] 
 \end{split}
\end{equation*}

\begin{equation*}
\begin{split}
\quad\quad\quad\quad =^{(c)}\mathbbm{E}\Bigg[\sum^T_{\tau=1}\Bigg(\max_i \frac{p_{t(\tau)+d_{t(\tau)}}(i)}{p_{t(\tau)}(i)}\Bigg)\times \\
\mathbbm{E}_{\Vec{a}_{t(\tau)}}\Bigg[\Vec{p}^\mathbb{T}_{t(\tau)}\hat{\vec{l}}_{t(\tau)|t(\tau)+d_{t(\tau)}}-\Vec{p}^\mathbb{T}\hat{\Vec{l}}_{t(\tau)|t(\tau)+d_{t(\tau)}}\Bigg]\Bigg]
\end{split}
\end{equation*}

\begin{equation*}
=^{(d)}\mathbbm{E}\Bigg[\sum^T_{\tau=1}\Bigg(\max_i \frac{p_{t(\tau)+d_{t(\tau)}}(i)}{p_{t(\tau)}(i)}\Bigg)\mathbbm{E}_{\Vec{a}_{t(\tau)}}\Bigg[\Vec{p}^\mathbb{T}_{t(\tau)}\tilde{\Vec{l}}_\tau  - \Vec{p}^\mathbb{T}\tilde{\Vec{l}}_\tau\Bigg]\Bigg]    
\end{equation*}

\begin{equation*}
=^{(e)}\mathbbm{E}\Bigg[\sum^T_{t=1}\Bigg(\max_i \frac{p_{t(\tau)+d_{t(\tau)}}(i)}{p_{t(\tau)}(i)}\Bigg)\mathbbm{E}_{\Vec{a}_{t(\tau)}}\Bigg[\tilde{\vec{p}}^\mathbb{T}_{\tau-\tilde{s}_\tau}\tilde{\vec{l}}_\tau - \Vec{p}^\mathbb{T}\tilde{\Vec{l}}_\tau\Bigg]\Bigg]    
\end{equation*}

\begin{equation}
\begin{split}
 =\mathbbm{E}\Bigg[\sum^T_{\tau=1}\Bigg(\max_i \frac{p_{t(\tau)+d_{t(\tau)}}(i)}{p_{t(\tau)}(i)}\Bigg)\Bigg(\mathbbm{E}_{\vec{a}_{t(\tau)}}\Bigg[\tilde{\Vec{p}}^\mathbb{T}_{\tau - \tilde{s}_\tau}\tilde{\Vec{l}}_\tau - \tilde{\Vec{p}}^\mathbb{T}_\tau \tilde{\vec{l}}_\tau\Bigg]+ \\ \Bigg(\mathbbm{E}_{\vec{a}_{t(\tau)}}\Bigg[\tilde{\Vec{p}}^\mathbb{T}_\tau \tilde{\Vec{l}}_\tau - \Vec{p}^\mathbb{T} \tilde{\vec{l}}_\tau\Bigg] \Bigg)
\label{eqn 43}
\end{split}
\end{equation}

(c) comes from the knowledge that $\{t(1),...,t(T)\}$ is a permutation of $\{1,...,T\}$; (d) results from $\tilde{\Vec{l}}_\tau = \hat{\Vec{l}}_{t(\tau)|t(\tau)+d_{t(\tau)}}$ and (e) is obtained from the (\ref{eqn 40}), so $\Vec{p}_{t(\tau)}=\tilde{\Vec{p}}_{L_{t(\tau)-1}+1}=\tilde{\Vec{p}}_{\tau - \tilde{s}_\tau}$. The rest of the proof is omitted due to space constraints.

\section{Conclusion}
In conclusion, the regret of the multiple play version of delayed adversarial bandit is only $\sqrt{k}$ times worse than the regret bound obtained in \cite{li2019bandit}. The multiple arms chosen each time are assumed to experience the same delay. 

\bibliographystyle{plainnat}
\bibliography{Ref}
\end{document}